\RequirePackage[T1]{fontenc}
\documentclass[letterpaper,10pt,conference]{ieeeconf}
\IEEEoverridecommandlockouts
\usepackage{mathptmx}
\usepackage{graphicx}
\usepackage{amsmath,amssymb}
\usepackage{booktabs}
\usepackage{cite}
\usepackage{url}
\usepackage[hidelinks]{hyperref}
\hypersetup{
  pdftitle={DAPEVO: Deep Adaptive Patch Frame-Event Visual Odometry},
  pdfauthor={Luca Gandolfi, Simone Nascivera, Roberto Pellerito, Rong Zou, Chiara Plizzari, Davide Scaramuzza},
  pdfsubject={Visual odometry with frames and event cameras},
  pdfkeywords={Visual odometry, Event cameras, Sensor fusion, Deep learning, Robotics}
}
\usepackage{tikz}
\usetikzlibrary{arrows.meta,positioning,calc}
\graphicspath{{figures/}}

\definecolor{resultbest}{HTML}{E2F0D7}
\definecolor{resultsecond}{HTML}{FFEFE3}
\definecolor{resultmethod}{HTML}{EDF1F7}

\title{\LARGE\bf DAPEVO: Deep Adaptive Patch Frame-Event \\ Visual Odometry}
\author{Luca Gandolfi$^{1,2}$, Simone Nascivera$^{1}$, Roberto Pellerito$^{1}$,\\
Rong Zou$^{1}$, Chiara Plizzari$^{2}$, and Davide Scaramuzza$^{1}$%
\thanks{This work was supported by the European Space Agency under Cooperative Agreement No. 4000150780/25/NL/GLC (with co-funding from Thales Alenia Space France), the European Union's Horizon Europe Research and Innovation Programme under grant agreement No. 101120732 (AUTOASSESS), and the European Research Council (ERC) under grant agreement No. 864042 (AGILEFLIGHT). The views expressed in this publication can in no way be taken to reflect the official opinion of the European Space Agency.}%
\thanks{$^{1}$Luca Gandolfi, Simone Nascivera, Roberto Pellerito, Rong Zou, and Davide Scaramuzza are with the Robotics and Perception Group, University of Zurich, Switzerland (\url{https://rpg.ifi.uzh.ch/})}%
\thanks{$^{2}$Luca Gandolfi and Chiara Plizzari are with Bocconi University, Italy.}%
}
\date{}
\begin{document}
\maketitle
\thispagestyle{empty}
\pagestyle{empty}
\suppressfloats[t]

\begin{abstract}
Visual odometry is essential for autonomous navigation in GPS-denied environments, yet RGB-based methods remain vulnerable to motion blur, challenging illumination, and dropped frames.
Event cameras complement conventional cameras with high temporal resolution and dynamic range, but their asynchronous measurements complicate reliable correspondence estimation.
We present DAPEVO, a learned visual odometry system that estimates image and event correspondences independently at shared patch locations and fuses their correlation evidence before motion refinement.
Each tracked patch maintains image and event descriptors, and a learned scalar gate combines modality-specific correlation embeddings for each patch--frame edge before a shared recurrent refinement and bundle-adjustment update.
DAPEVO also supports event-only observations, enabling continued tracking when RGB frames are sparse or unavailable, while modality-aware keyframe culling preserves scarce frame constraints.
On UZH-FPV, when retaining only one in six RGB frames, DAPEVO's mean absolute trajectory error (ATE) increases by only 36\%, from 1.00 to 1.36\,m, whereas the ATE of DPVO and RAMP-VO rises by factors of $3.7\times$ and $3.1\times$, respectively.
On TartanEvent, DAPEVO similarly remains below 1\,m ATE at 3\,Hz RGB input, while DPVO and RAMP-VO exceed 9\,m.
Under degraded RGB input on TartanEvent, DAPEVO achieves an ATE of 0.60\,m, compared with more than 4\,m for both DPVO and RAMP-VO, while also outperforming event-only DEVO at 0.87\,m.
\end{abstract}

\section{Introduction}
\label{sec:intro}

\begin{figure}[t]
  \centering
  \includegraphics[width=\columnwidth]{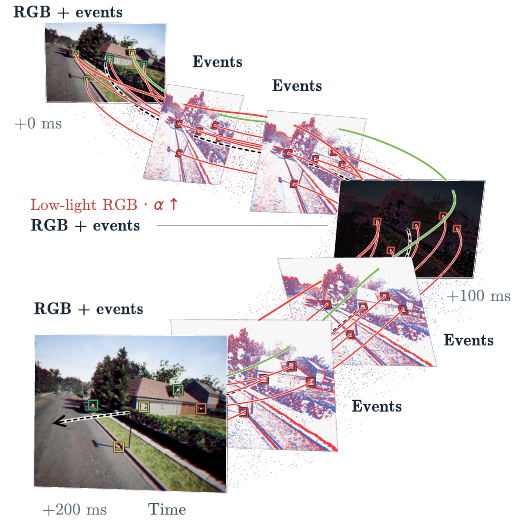}
  \caption{\textbf{Overview of DAPEVO.}
Joint RGB–event and event-only observations are processed by separate image and event encoders.
Patches with RGB and event descriptors are correlated independently, then adaptively fused by a learned gate before recurrent refinement and bundle adjustment.
Patch colors and track lines indicate the event contribution $\alpha$, from green ($\alpha=0$) to red ($\alpha=1$).
A low-light augmentation is applied to the central RGB observation to illustrate reduced reliance on degraded frame information.
When RGB is unavailable at either correspondence endpoint, $\alpha=1$, enabling event-only tracking.}

  \label{fig:DAPEVO}
\end{figure}

Reliable localization in GPS-denied environments is essential for autonomous robots operating indoors, underground, and in planetary exploration settings~\cite{carlone2026slam,tranzatto2024team,zhao2024subt,cheng2005visual}.
Image-based visual odometry provides rich appearance information, but motion blur, limited dynamic range, and sparse or dropped images can degrade the correspondences required for motion estimation.
Event cameras complement conventional cameras with high temporal resolution and high dynamic range~\cite{gallego2022survey}, providing measurements between image acquisitions and under conditions in which RGB-based tracking becomes unreliable.

The relative usefulness of the two modalities varies spatially and temporally.
A visually distinctive image region may generate few informative events, while degraded image observations may coincide with strong event measurements.
A robust image-event estimator must therefore adapt to both the availability and reliability of each modality.

DPVO~\cite{teed2023deep} combines sparse patch tracking with learned correspondence refinement and differentiable bundle adjustment, while DEVO~\cite{klenk2024devo} adapts this formulation to event data.
RGB--event methods such as EDS~\cite{hidalgo2022eds} and RAMP-VO~\cite{pellerito2024rampvo} exploit the complementary sensing modalities.
In particular, RAMP-VO asynchronously combines incoming image and event measurements into a shared recurrent feature representation from which matching and context features are decoded.
This shared representation enables multimodal estimation, but does not explicitly preserve and weight modality-specific evidence.

This distinction becomes important when one modality becomes unreliable, as its features can introduce ambiguous correspondences and noisy constraints into the patch graph.
Event measurements pose an additional challenge because their differential and motion-dependent nature makes correspondence estimation more difficult than conventional image matching.

We introduce Deep Adaptive Patch Frame-Event Visual Odometry (DAPEVO), a monocular RGB-event odometry method built on DPVO.
Co-located image and event descriptors are matched independently, and their correlation embeddings are fused by an edge- and iteration-specific learned gate.
When RGB is unavailable, event-only observations remain in the same optimization graph, while modality-aware keyframe culling preserves sparse RGB constraints.

DAPEVO remains competitive under nominal conditions while providing larger gains when RGB observations become sparse or degraded.
On UZH-FPV~\cite{delmerico2019uzh}, when retaining only one in six RGB frames, DAPEVO increases its ATE by only 36\%, whereas the ATE of DPVO and RAMP-VO rises by factors of $3.7\times$ and $3.1\times$, respectively.
On TartanEvent~\cite{pellerito2024rampvo}, DAPEVO remains below 1\,m ATE at 3\,Hz RGB input, while DPVO and RAMP-VO exceed 9\,m.
Under degraded RGB input on TartanEvent, DAPEVO achieves an ATE of 0.60\,m, compared with more than 4\,m for both DPVO and RAMP-VO, while also outperforming event-only DEVO at 0.87\,m.

Our contributions are:
\begin{itemize}
    \item A correspondence-level RGB-event fusion strategy that preserves modality-specific matching evidence and adaptively combines it for each patch--frame edge and recurrent refinement iteration.
    \item Joint RGB--event and event-only observations within the same optimization graph, together with modality-aware keyframe culling that preserves sparse RGB-bearing constraints, enabling patch tracking and pose updates at a higher rate than the RGB frame rate.
    \item An evaluation on synthetic and real-world datasets under nominal operation, degraded RGB observations, and reduced RGB frame rates.
\end{itemize}

\begin{figure*}[!t]
  \centering
  \includegraphics[width=\textwidth]{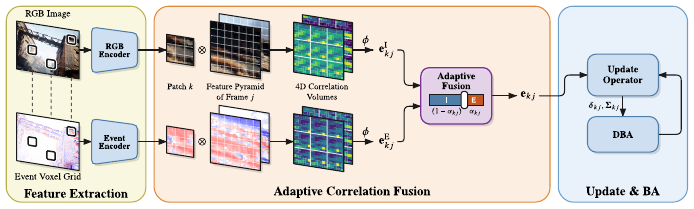}
\caption{\textbf{DAPEVO architecture.} Images and event voxel grids are processed by separate matching encoders. Each patch $k$ is associated with image and event matching features $\mathbf g_k^I$ and $\mathbf g_k^E$. For each patch--frame edge $(k,j)$, image and event correlations are computed independently and mapped to embeddings $\mathbf e^I_{kj}$ and $\mathbf e^E_{kj}$. A learned gate predicts an edge-specific weight $\alpha_{kj}$ to fuse the two embeddings. When RGB images are unavailable, only the event embedding is used. The resulting embedding $\mathbf e_{kj}$ is processed by the recurrent update operator to predict correspondence corrections and confidence weights, which are then used by differentiable bundle adjustment (DBA) to jointly refine camera poses and patch depths.}

  \label{fig:method}
\end{figure*}

\section{Related Work}
\label{sec:related}
\subsection{Geometric and Learned Visual Odometry}
Classical visual odometry estimates geometric constraints through feature matching, as in ORB-SLAM3~\cite{campos2021orbslam3}, or direct photometric alignment, as in DSO~\cite{engel2018dso}.
DeepVO~\cite{wang2017deepvo} regresses camera motion with a recurrent network, whereas D3VO~\cite{yang2020d3vo} integrates predicted depth, pose, and photometric uncertainty into a direct geometric estimator.
RAFT~\cite{teed2020raft} introduced recurrent refinement through correlation lookups for optical flow.
DROID-SLAM~\cite{teed2021droid} uses recurrent correspondence refinement and differentiable bundle adjustment, while DPVO~\cite{teed2023deep} introduces a sparse patch representation for efficient visual odometry. 
DPV-SLAM~\cite{lipson2024deep} extends this formulation with loop closure.

\subsection{Event-Based Motion Estimation}
Early event-based estimators recover motion and structure through filtering or geometric optimization.
Kim et al.~\cite{kim2016tracking} jointly estimate motion and structure from monocular events, while EVO~\cite{rebecq2017evo} performs event-to-map alignment with parallel mapping.
ESVO~\cite{zhou2021esvo} extends event-based odometry to stereo, and event-inertial approaches combine event tracks with filtering or nonlinear optimization~\cite{zhu2017evio,rebecq2017eventvio}.

Learned methods estimate motion directly from event representations or integrate learned correspondence estimation with geometric optimization.
Zhu et al.~\cite{zhu2019unsupervised} jointly learn optical flow, depth, and egomotion through event motion compensation.
DEVO~\cite{klenk2024devo} adapts DPVO to event voxel grids and introduces learned event-based patch selection, while DEIO~\cite{deio} and Stereo-DEVO~\cite{stereodevo} extend learned event odometry with inertial and stereo measurements.

\subsection{Combining Frames and Events}
RGB--event methods exploit the complementary appearance and temporal information provided by the two sensing modalities.
Kueng et al.~\cite{kueng2016tracks} detect features in intensity images and track them asynchronously using events.
EDS~\cite{hidalgo2022eds} estimates inter-frame motion from an event-derived brightness-increment residual and explicitly evaluates reduced frame rates.
Ultimate SLAM~\cite{vidal2018ultimate} and PL-EVIO~\cite{guan2023plevio} combine frame, event, and inertial constraints within geometric estimators.

Learned approaches have also explored recurrent and shared representations for asynchronous multimodal fusion.
RAM-Net~\cite{gehrig2021ramnet} maintains a recurrent state that can be updated asynchronously by either modality.
RAMP-VO~\cite{pellerito2024rampvo} extends this formulation to visual odometry using parallel multiscale encoders that integrate frame and event inputs into a shared recurrent representation, from which matching and context features are decoded.
It supports asynchronous inputs and reduced RGB frame rates, but does not retain separate image- and event-based correlation evidence through correspondence estimation.

\section{Method}
\label{sec:method}
\subsection{Input Representation and Geometric State}

Figure~\ref{fig:method} provides an overview of the DAPEVO architecture. At timestamp $t$, DAPEVO receives a five-bin event voxel grid $\mathbf V_t\in\mathbb{R}^{5\times H\times W}$ and, when available, an RGB image $\mathbf I_t\in\mathbb{R}^{3\times H\times W}$. We refer to observations containing both modalities as \emph{joint frames} and those containing only events as \emph{event-only frames}. Both modalities are assumed to share the same camera intrinsics, as provided by a shared pixel array or a calibrated beamsplitter configuration. Following DEVO~\cite{klenk2024devo}, events are accumulated into five temporal bins; in contrast to the RGB inputs, \emph{event-only} frames can be instantiated at arbitrary timestamps.

We adopt the patch representation and geometric formulation of DPVO~\cite{teed2023deep}, together with the event-based patch selection of DEVO~\cite{klenk2024devo}. The geometric state consists of $N$ camera poses $\mathbf T\in\mathrm{SE}(3)^N$, one for each frame, and a set of patches $\{\mathbf P_k\}_k$, with a fixed number of patches extracted from each incoming frame. Each patch $\mathbf P_k\in\mathbb{R}^{4\times p^2}$ represents a $p\times p$ region in homogeneous coordinates, with its fourth row containing a constant inverse depth $d_k$ shared by all patch elements. As in DPVO, we use $p=3$.

Patches and frames form a bipartite patch graph, with an edge $(k,j)\in\mathcal E$ connecting patch $k$, extracted from frame $i$, to a target frame $j$. As in DPVO~\cite{teed2023deep}, patch $k$ is reprojected from its source frame $i$ onto frame $j$ according to
\begin{equation}
    \mathbf P'_{kj} = \omega_{ij}(\mathbf T,\mathbf P_k),
    \label{eq:projection}
\end{equation}
where $\omega_{ij}$ denotes the geometric reprojection from frame $i$ to frame $j$.

\subsection{Feature Extraction}
RGB images and event voxel grids are processed by separate matching and context encoders. The matching encoders produce 128-channel feature maps at $1/4$ of the input resolution. A second level at $1/16$ resolution is obtained by average pooling, forming a two-level matching feature pyramid for each modality that provides the target-frame features used for correlation.
The matching features of patch $k$ are cropped from the corresponding $1/4$-resolution feature maps of its source frame $i$ using bilinear sampling. For a patch $k$ extracted from a joint frame, this yields modality-specific matching features
\begin{equation}
\mathbf g_k^I,\mathbf g_k^E \in \mathbb{R}^{128\times p\times p},
\end{equation}
where $\mathbf g_k^I$ and $\mathbf g_k^E$ denote the image and event matching features, respectively.
For a patch extracted from an event-only frame, only $\mathbf g_k^E$ is available.

The context encoders produce 384-channel feature maps at $1/4$ of the input resolution. For each patch $k$ extracted from frame $i$, an event context feature $\mathbf u_k^E\in\mathbb{R}^{384}$ is sampled at the patch center from the event context feature map of frame $i$. If $i$ is joint, an image context feature $\mathbf u_k^I\in\mathbb{R}^{384}$ is sampled at the same location from the corresponding image context feature map. The context feature associated with patch $k$ is then defined as
\begin{equation}
\mathbf u_k =
\begin{cases}
\mathbf u_k^I + \psi\!\left([\mathbf u_k^I;\mathbf u_k^E]\right),
& \text{if $i$ is joint}, \\
\mathbf u_k^E,
& \text{if $i$ is event-only},
\end{cases}
\label{eq:context}
\end{equation}
where $[\cdot;\cdot]$ denotes concatenation and $\psi$ is a two-layer multilayer perceptron (MLP) with a ReLU nonlinearity. When $i$ is joint, $\mathbf u_k^I$ provides the base representation, while event information is incorporated through a learned residual. The resulting context feature $\mathbf u_k$ is computed once when the patch is instantiated and shared across all edges associated with that patch.

\subsection{Adaptive Correlation Fusion}
Following DPVO~\cite{teed2023deep} and DEVO~\cite{klenk2024devo}, each pixel of patch $k$ is correlated with a $7\times7$ neighborhood centered at its current reprojection in the target frame $j$, independently at both levels of the corresponding matching feature pyramid, as illustrated in Fig.~\ref{fig:method}. This produces a $p\times p\times7\times7$ correlation volume at each pyramid level and for each modality. The two levels are flattened and concatenated, yielding
\begin{equation}
\mathbf c_{kj}^I,\mathbf c_{kj}^E
\in\mathbb{R}^{2p^2 7^2}
=\mathbb{R}^{882}.
\end{equation}
While image and event features remain in separate modality-specific feature spaces, their correlation volumes provide a common representation of matching evidence. We therefore use a shared correlation encoder $\phi$ to map the two correlation vectors to modality-specific embeddings,
\begin{equation}
\mathbf e_{kj}^I=\phi(\mathbf c_{kj}^I),
\qquad
\mathbf e_{kj}^E=\phi(\mathbf c_{kj}^E),
\qquad
\mathbf e_{kj}^I,\mathbf e_{kj}^E\in\mathbb{R}^{384}.
\label{eq:embedding}
\end{equation}
An edge-specific fusion weight $\alpha_{kj}\in[0,1]$ is predicted from the two embeddings as
\begin{equation}
\alpha_{kj}
=
\gamma\!\left([\mathbf e_{kj}^I;\mathbf e_{kj}^E]\right),
\label{eq:gate}
\end{equation}
where $\gamma$ denotes a learned linear layer followed by a sigmoid activation.

If either $i$ or $j$ is event-only, image matching is unavailable and only the event correlation embedding $\mathbf e_{kj}^E$ is computed. The correlation embedding used by the recurrent update operator is therefore
\begin{equation}
\mathbf e_{kj}
=
\begin{cases}
(1-\alpha_{kj})\mathbf e_{kj}^I
+\alpha_{kj}\mathbf e_{kj}^E,
& \text{if $i$ and $j$ are joint}, \\
\mathbf e_{kj}^E,
& \text{otherwise}.
\end{cases}
\label{eq:fusion}
\end{equation}
Since $\alpha_{kj}$ is recomputed at each recurrent iteration, the fusion adapts independently for each edge and refinement step.

\subsection{Update Operator and Differentiable Bundle Adjustment}

Following DPVO~\cite{teed2023deep} and DEVO~\cite{klenk2024devo}, correspondence refinement and geometric optimization are performed by the recurrent update operator and differentiable bundle adjustment.
DAPEVO retains these components unchanged, replacing the single-modality correlation input with the fused embedding $\mathbf e_{kj}$ from~(\ref{eq:fusion}).
The update operator predicts 2D flow revisions $\{\boldsymbol\delta_{kj}\}_{kj}$ and confidence weights $\{\boldsymbol\Sigma_{kj}\}_{kj}$, which the bundle adjustment uses to jointly refine camera poses and patch depths.

\subsection{Event-Only Tracking and Keyframe Management}
\label{sec:tracking}

DAPEVO supports event-only tracking between RGB observations, allowing the tracking rate to exceed the RGB frame rate. During initialization, only joint frames are used, while incoming event-only frames are discarded. After initialization, both joint and event-only frames are processed.

To preserve joint observations when RGB frames are sparse, we introduce a modality-aware variant of the DPVO~\cite{teed2023deep} keyframe-culling criterion, which we refer to as \emph{subgraph culling}. Projected motion is evaluated with respect to the nearest preceding and following frames that share the modalities available in the candidate. For a joint candidate, the nearest joint frames are used, whereas for an event-only candidate, the nearest frames are used regardless of type, since event information is available in every frame. The candidate is removed when the mean bidirectional projected motion falls below 15 feature-map pixels for joint frames or 25 pixels for event-only frames. These thresholds should be adjusted when using different input resolutions.

\subsection{Training}
\label{sec:training}

DAPEVO is trained in two stages using the supervision and recurrent optimization framework of DPVO~\cite{teed2023deep}.
In the first stage, the event matching and context encoders and the event-based patch scorer are trained while the pretrained DPVO update operator remains frozen.
This stage learns event representations that are compatible with the update operator originally trained on image features.

In the second stage, the pretrained image branch is combined with the learned event branch and the two feature extractors are frozen.
The context-fusion MLP $\psi$ and correlation gate $\gamma$ are then optimized to combine the modality-specific representations.
The recurrent update operator is subsequently fine-tuned jointly with the fusion modules.

To prevent the fusion mechanism from collapsing to the generally stronger image representation under nominal conditions, training includes modality-specific input corruption and modality dropout.
Image corruption comprises low-light augmentation, noise, blur, and occlusion, whereas event corruption comprises temporal dropout, event thinning, and noise.
Modality dropout additionally removes the correlation evidence of either modality, exposing the recurrent update operator to single-modality observations during training.

\section{Experiments}
\label{sec:experiments}

\subsection{Implementation Details}
\label{sec:implementation}

DAPEVO is trained on the TartanEvent training set \cite{tartanair, pellerito2024rampvo}. The event branch is trained for 100,000 iterations, with the first 1,000 iterations using the structure-only warm-up of DPVO~\cite{teed2023deep}.
The subsequent fusion stage is trained for 60,000 iterations.
The final layer of the context-fusion MLP $\psi$ is initialized to zero, while the correlation gate $\gamma$ is initialized with zero weights and bias $-2$, yielding $\alpha_{kj}\simeq0.12$ and an initially image-dominant representation.

During fusion training, image corruption is applied to 50\% of sequences and event corruption to 15\%, while the remaining 35\% are left unmodified.
Image and event correlation volumes are independently dropped in 5\% of training sequences for each modality.
After 8,000 fusion-training iterations, the recurrent update operator is unfrozen and optimized with a learning rate ten times lower than that of the fusion modules.

The network is trained on a single RTX8000 GPU using the same training hyperparameters from DPVO and DEVO. All experiments use the default DPVO configuration~\cite{teed2023deep}, with 96 patches extracted per input frame.

In addition, we optimize the runtime of the existing DPVO codebase, without changing its network structure.
Batched aggregation over patch and frame-pair groups replaces generic scatter operations and reuses indexing layouts between graph changes.
Static half precision and compilation reduce conversion and kernel-launch costs.
The correlation kernel stages overlapping neighborhoods in shared memory and uses channels-last vectorized feature loads.

Using its default configuration, these optimizations reduce DPVO processing time on the Quadro RTX 8000 from 55.8 to 23.8\,ms per input frame. Under the same configuration, DAPEVO requires 36.5\,ms per frame when every input contains both RGB and events, corresponding to a 53\% overhead over optimized DPVO. On an RTX 5090, optimized DPVO and DAPEVO require 7.5 and 10.2\,ms per frame, respectively, corresponding to a 36\% overhead.

\subsection{Experimental Setup}
\label{sec:experimental_setup}

We evaluate DAPEVO on synthetic and real-world event-camera benchmarks with synchronized RGB images and events.
TartanEvent~\cite{pellerito2024rampvo} provides synthetic events, images, and ground-truth trajectories generated from TartanAir~\cite{tartanair}; we use its 32-sequence validation split and the monocular test split of the ECCV 2020 SLAM Competition~\cite{tartanair, pellerito2024rampvo}.
We additionally evaluate on UZH-FPV~\cite{delmerico2019uzh} for aggressive quadrotor motion, the HKU monocular~\cite{guan2023plevio,guan2022monocular} 
sequences under challenging illumination and high-dynamic-range conditions, and the four indoor-flying sequences of MVSEC~\cite{zhu2018multivehicle}.

We report absolute trajectory error (ATE) after $\mathrm{Sim}(3)$ alignment on TartanEvent and UZH-FPV.
For TartanEvent, we additionally report the area under the empirical cumulative distribution of sequence ATE over $[0,1]$ m (AUC).
On HKU, UZH FPV and MVSEC, we report mean position error (MPE), defined as the mean position error normalized by the ground-truth trajectory length, expressed as a percentage.
Following PL-EVIO~\cite{guan2023plevio}, HKU trajectories are aligned using the first $5$ s, and the resulting transformation is held fixed for the remainder of the trajectory.
For DPVO, DEVO, RAMP-VO and DAPEVO, this alignment additionally estimates the global scale.
Additionally, in the following tables \textit{fail} denotes a failure explicitly reported by the original source. 
The cited evaluations do not specify a numerical failure threshold.

\subsection{Benchmark Comparison}
\label{sec:benchmark}
\textbf{TartanEvent}.
Under nominal RGB input, DAPEVO and RAMP-VO achieve identical ATE and AUC at the reported precision (Table~\ref{tab:tartan_dark}).
This is consistent with evaluation in the noise-free synthetic training domain, where reliable image correspondences may limit the benefit of adaptive fusion.
To isolate the fusion architecture, we train a \textit{Feature Fusion} variant using the same data, augmentations, and optimization protocol as DAPEVO.
It concatenates image and event features from separate convolutional stems, then uses a shared encoder to produce joint patch descriptors and a single correlation volume.
Its ATE is $0.22$ m under nominal RGB input and $1.68$ m under degradation, compared with $0.15$ m and $0.60$ m for DAPEVO.
The $64.3\%$ reduction under degradation supports the benefit of separate correlations and adaptive fusion beyond the training protocol; Sec.~\ref{sec:rgb_degradation} examines this condition further.
\begin{table}[t]
    \centering
    \caption{TartanEvent validation results under nominal and degraded RGB input.}
    \label{tab:tartan_dark}
    \scriptsize
    \setlength{\tabcolsep}{5pt}
    \begin{tabular}{lc|cc|cc}
        \toprule
        & & \multicolumn{2}{c|}{Nominal RGB}
          & \multicolumn{2}{c}{Degraded RGB} \\
        Method & Input
        & AUC $\uparrow$ & ATE [m] $\downarrow$
        & AUC $\uparrow$ & ATE [m] $\downarrow$ \\
        \midrule
        DPVO~\cite{teed2023deep} & I
        & 0.83 & 0.17
        & 0.14 & 4.96 \\
        DEVO~\cite{klenk2024devo} & E
        & 0.56 & 0.87
        & 0.56 & 0.87 \\
        RAMP-VO~\cite{pellerito2024rampvo} & I+E
        & \textbf{0.85} & \textbf{0.15}
        & 0.14 & 7.01 \\
        Feature fusion & I+E
        & 0.78 & 0.22
        & 0.50 & 1.68 \\
        DAPEVO   & I+E
        & \textbf{0.85} & \textbf{0.15}
        & \textbf{0.63} & \textbf{0.60} \\
        \bottomrule
    \end{tabular}
\end{table}

\textbf{UZH-FPV}. DAPEVO achieves the lowest average MPE on UZH-FPV, reducing the error to $0.46$ from $0.52$ for DPVO and $0.53$ for DEVO (Table~\ref{tab:uzh_fpv_mpe}).
For comparability with published results, Table~\ref{tab:uzh_fpv_mpe} uses the sequence subset adopted in prior evaluations; the reduced-frame-rate experiments in Sec.~\ref{sec:rgb_degradation} instead evaluate all UZH-FPV sequences.
\begin{table}[t]
    \centering
    \caption{
        MPE [\%] on indoor UZH-FPV.
        All results are from \cite{guan2023plevio,klenk2024devo}, except for DPVO and DAPEVO. Best results are bold.
    }
    \label{tab:uzh_fpv_mpe}
    \scriptsize
    \setlength{\tabcolsep}{2.2pt}
    \begin{tabular}{lcccccc}
        \toprule
        & \shortstack{VINS-\\Mono\\\cite{qin2018vinsmono}}
        & \shortstack{Ultimate\\SLAM\\\cite{vidal2018ultimate}}
        & \shortstack{PL-EVIO\\\cite{guan2023plevio}}
        & \shortstack{DEVO\\\cite{klenk2024devo}}
        & \shortstack{DPVO\\\cite{teed2023deep}}
        & \shortstack{DAPEVO\\\strut} \\
        \shortstack[l]{Loop\\closure}
        & \checkmark
        & \checkmark
        & \checkmark
        & $\times$
        & $\times$
        & $\times$ \\
        \midrule
        fwd\_3
        & 0.65 & \textit{fail} & 0.38 & 0.37
        & \textbf{0.32} & 0.38 \\
        fwd\_5
        & 1.07 & \textit{fail} & 0.90 & 0.40
        & \textbf{0.24} & 0.25 \\
        fwd\_6
        & \textbf{0.25} & \textit{fail} & 0.30 & 0.31
        & 0.42 & 0.39 \\
        fwd\_7
        & 0.37 & \textit{fail} & 0.55 & 0.50
        & 0.23 & \textbf{0.19} \\
        fwd\_9
        & 0.51 & \textit{fail} & \textbf{0.44} & 0.61
        & 0.57 & 0.56 \\
        fwd\_10
        & 0.92 & \textit{fail} & 1.06 & \textbf{0.52}
        & 0.58 & 0.54 \\
        45deg\_2
        & \textbf{0.53} & \textit{fail} & 0.55 & 0.72
        & 0.81 & 0.69 \\
        45deg\_4
        & 1.72 & 9.79 & 1.30 & 0.45
        & 0.62 & \textbf{0.43} \\
        45deg\_9
        & 1.25 & 4.74 & 0.76 & 0.89
        & 0.93 & \textbf{0.73} \\
        \midrule
        Avg.
        & 0.81
        & fail
        & 0.69
        & 0.53
        & 0.52
        & \textbf{0.46} \\
        \bottomrule
    \end{tabular}
\end{table}

\textbf{HKU Monocular}. The HKU monocular sequences expose a different behavior, as summarized in Table~\ref{tab:hku_mpe}.
DAPEVO and DPVO have similar median errors across trials, with average median MPEs of $0.19$ and $0.21$  respectively, but their mean errors differ markedly: $0.19$ for DAPEVO versus $2.32$ for DPVO.
The discrepancy is dominated by occasional high-error DPVO trials, most notably on \texttt{d2l2}, where its mean/median MPE reaches $22.55/0.30$  compared with $0.19/0.19$ for DAPEVO.
DAPEVO therefore maintains comparable nominal accuracy while exhibiting substantially greater consistency across repeated trials under the challenging illumination conditions represented in this benchmark.
\begin{table}[t]
    \centering
    \caption{
        HKU monocular MPE [\%]. DPVO/DAPEVO report mean/median over five trials;
        other results are from~\cite{guan2023plevio} and use an IMU.
        I: images; E: events.
    }
    \label{tab:hku_mpe}
    \scriptsize
    \setlength{\tabcolsep}{1.6pt}
    \begin{tabular}{@{}lccccc@{}}
        \toprule
        & \shortstack{Ultimate\\SLAM\\\cite{vidal2018ultimate}}
        & \shortstack{PL-EIO\\\cite{guan2023plevio}}
        & \shortstack{PL-EVIO\\\cite{guan2023plevio}}
        & \shortstack{DPVO\\\cite{teed2023deep}}
        & \shortstack{DAPEVO\\\strut} \\
        & & & &
        \multicolumn{2}{c}{Mean / Median} \\
        \midrule
        Input
        & I+E
        & E
        & I+E
        & I
        & I+E \\
        Loop clos.
        & \checkmark
        & \checkmark
        & \checkmark
        & $\times$
        & $\times$ \\
        \midrule
        hdr1
        & 2.44 & 0.67 & 0.17
        & \textbf{0.11 / 0.12} & 0.12 / 0.12 \\
        hdr2
        & 1.11 & 0.45 & \textbf{0.12}
        & 0.17 / 0.16 & 0.15 / 0.15 \\
        hdr3
        & 0.83 & 0.74 & \textbf{0.19}
        & 0.27 / 0.27 & 0.26 / 0.25 \\
        hdr4
        & 1.49 & 0.37 & \textbf{0.11}
        & 0.18 / 0.18 & 0.15 / 0.16 \\
        d2l1
        & 1.00 & 0.78 & \textbf{0.14}
        & 0.20 / 0.20 & 0.21 / 0.21 \\
        d2l2
        & 0.79 & 0.44 & \textbf{0.12}
        & 22.55 / 0.30 & 0.19 / 0.19 \\
        l2d1
        & 0.84 & 0.42 & \textbf{0.13}
        & 0.90 / 0.31 & 0.28 / 0.28 \\
        l2d2
        & 1.49 & 0.73 & \textbf{0.16}
        & 0.17 / 0.16 & \textbf{0.16 / 0.16} \\
        dark1
        & 3.45 & 0.64 & 0.43
        & \textbf{0.12 / 0.12} & \textbf{0.12 / 0.12} \\
        dark2
        & 0.63 & 0.30 & 0.47
        & \textbf{0.20 / 0.19} & \textbf{0.20 / 0.21} \\
        aggr.\ hdr
        & 2.30 & 0.62 & 1.97
        & 0.65 / 0.28 & \textbf{0.21 / 0.19} \\
        \midrule
        Avg.
        & 1.49
        & 0.56
        & 0.36
        & 2.32 / 0.21
        & \textbf{0.19 / 0.19} \\
        \bottomrule
    \end{tabular}
\end{table}

\textbf{MVSEC}. DAPEVO achieves an average MPE of $0.19\%$, close to RGB-only DPVO at $0.17\%$ and below event-only DEVO at $0.46\%$ (Table~\ref{tab:mvsec_mpe}).
The evaluated learned methods outperform the classical baselines, and incorporating events preserves comparable RGB frame-based accuracy.
\begin{table}[t]
    \centering
    \caption{
        MVSEC indoor-flying MPE [\%]. Best/second-best results are bold/underlined.
    }
    \label{tab:mvsec_mpe}
    \scriptsize
    \setlength{\tabcolsep}{1.5pt}
    \begin{tabular}{@{}lcccccccccc@{}}
        \toprule
        & \shortstack{ORB-\\SLAM3\\\cite{campos2021orbslam3}}
        & \shortstack{VINS-\\Fusion\\\cite{qin2019vinsfusion}}
        & \shortstack{ESVIO\\\cite{chen2023esvio}}
        & \shortstack{Ultimate\\SLAM\\\cite{vidal2018ultimate}}
        & \shortstack{PL-\\EVIO\\\cite{guan2023plevio}}
        & \shortstack{ESVO\\\cite{zhou2021esvo}}
        & \shortstack{EVO\\\cite{rebecq2017evo}}
        & \shortstack{DPVO\\\cite{teed2023deep}}
        & \shortstack{DEVO\\\cite{klenk2024devo}}
        & \shortstack{DAPEVO\\\strut} \\
        \midrule

        IF1
        & 5.31
        & 1.50
        & 0.94
        &fail
        & 1.35
        & 4.00
        & 5.09
        & \textbf{0.16}
        & 0.26
        & \textbf{0.16} \\

        IF2
        & 5.65
        & 6.98
        & 1.00
        &fail
        & 1.00
        & 3.66
        &fail
        & \textbf{0.15}
        & 0.32
        & \underline{0.16} \\

        IF3
        & 2.90
        & 0.73
        & 0.47
        &fail
        & 0.64
        & 1.71
        & 2.58
        & \textbf{0.08}
        & 0.19
        & \textbf{0.08} \\

        IF4
        & 6.99
        & 3.62
        & 5.55
        & 2.77
        & 5.31
        &fail
        &fail
        & \textbf{0.30}
        & 1.08
        & \underline{0.36} \\

        \midrule
        Avg.
        & 5.21
        & 3.21
        & 1.99
        &fail
        & 2.08
        &fail
        &fail
        & \textbf{0.17}
        & 0.46
        & \underline{0.19} \\

        \bottomrule
    \end{tabular}
\end{table}

\textbf{TartanEvent 2020 Competition}. Generalization to the held-out ECCV 2020 SLAM Competition split is evaluated in Table~\ref{tab:tartan_competition}.
DAPEVO achieves an average ATE of $0.19$ m, improving over DPVO at $0.21$ m but remaining behind RAMP-VO at $0.17$ m.
DAPEVO ranks second in average ATE and attains the best or tied-best result on seven of the sixteen sequences.
This evaluation provides an additional comparison on trajectories not contained in the TartanEvent validation split, but remains within the same noise-free synthetic domain used for training, where reliable RGB observations may limit the benefit of adaptive modality weighting.
\begin{table}[t]
    \centering
    \caption{
        ECCV 2020 SLAM Competition monocular with simulated event ATE [m]. DAPEVO: median of five trials;
        baselines:~\cite{pellerito2024rampvo}. Checkmarks denote global optimization or
        loop closure. Among other methods, best /second-best
        results are bold/underlined.
    }
    \label{tab:tartan_competition}
    \scriptsize
    \setlength{\tabcolsep}{1.4pt}
    \begin{tabular}{@{}lcccccccc@{}}
        \toprule
        & \shortstack{ORB-\\SLAM3\\\cite{campos2021orbslam3}}
        & \shortstack{COLMAP\\\cite{schoenberger2016sfm}}
        & \shortstack{DROID-\\SLAM\\\cite{teed2021droid}}
        & \shortstack{DSO\\\cite{engel2018dso}}
        & \shortstack{DROID-\\VO\\\cite{teed2021droid}}
        & \shortstack{DPVO\\\cite{teed2023deep}}
        & \shortstack{RAMP-\\VO\\\cite{pellerito2024rampvo}}
        & \shortstack{DAPEVO\\\strut} \\
        \midrule
        Input
        & I & I & I & I & I & I & I+E & I+E \\
        Loop clos.
        & \checkmark & \checkmark & \checkmark
        & $\times$ & $\times$ & $\times$ & $\times$ & $\times$ \\
        \midrule

        ME00
        & 13.61 & 15.20 & 0.17 & 9.65 & 0.22
        & \underline{0.16} & 0.20 & \textbf{0.14} \\

        ME01
        & 16.86 & 5.58 & 0.06 & 3.84 & 0.15
        & 0.11 & \textbf{0.04} & \underline{0.08} \\

        ME02
        & 20.57 & 10.86 & 0.36 & 12.20 & 0.24
        & \underline{0.11} & \textbf{0.10} & 0.14 \\

        ME03
        & 16.00 & 3.93 & 0.87 & 8.17 & 1.27
        & 0.66 & \underline{0.46} & \textbf{0.41} \\

        ME04
        & 22.27 & 2.62 & 1.14 & 9.27 & 1.04
        & \underline{0.31} & \textbf{0.16} & 0.35 \\

        ME05
        & 9.28 & 14.78 & 0.13 & 2.94 & 0.14
        & 0.14 & \underline{0.13} & \textbf{0.07} \\

        ME06
        & 21.61 & 7.00 & 1.13 & 8.15 & 1.32
        & 0.30 & \textbf{0.12} & \underline{0.21} \\

        ME07
        & 7.74 & 18.47 & 0.06 & 5.43 & 0.77
        & 0.13 & \underline{0.12} & \textbf{0.11} \\

        \midrule

        MH00
        & 15.44 & 12.26 & 0.08 & 9.92 & 0.32
        & \textbf{0.21} & 0.36 & \underline{0.28} \\

        MH01
        & 2.92 & 13.45 & 0.05 & 0.35 & 0.13
        & \underline{0.04} & 0.06 & \textbf{0.03} \\

        MH02
        & 13.51 & 13.45 & 0.04 & 7.96 & 0.08
        & \textbf{0.04} & \textbf{0.04} & \textbf{0.04} \\

        MH03
        & 8.18 & 20.95 & 0.02 & 3.46 & 0.09
        & 0.08 & \textbf{0.04} & \underline{0.05} \\

        MH04
        & 2.59 & 24.97 & 0.01 & fail
        & 1.52 & \underline{0.58} & \textbf{0.41} & 0.78 \\

        MH05
        & 21.91 & 16.79 & 0.68 & 12.58 & 0.69
        & \textbf{0.17} & 0.25 & \underline{0.21} \\

        MH06
        & 11.70 & 7.01 & 0.30 & 8.42 & 0.39
        & \underline{0.11} & \underline{0.11} & \textbf{0.10} \\

        MH07
        & 25.88 & 7.97 & 0.07 & 7.50 & 0.97
        & 0.15 & \textbf{0.07} & \underline{0.08} \\

        \midrule
        Avg.
        & 14.38 & 12.50 & 0.33 & fail & 0.58
        & 0.21 & \textbf{0.17} & \underline{0.19} \\
        \bottomrule
    \end{tabular}
\end{table}

\subsection{Robustness to RGB Degradation and Reduced Frame Rate}
\label{sec:rgb_degradation}

We evaluate DAPEVO under two conditions that reduce the reliability or availability of image-based correspondences: degraded RGB observations and reduced RGB frame rates.

\textbf{Degraded RGB observations.}
We construct a degraded variant of the TartanEvent validation set, denoted \emph{TartanDark}, by applying darkening gamma correction and adding gaussian noise to every image while leaving the event stream unchanged.
All methods are evaluated on the same degraded sequences, and results are averaged over five trials.
Table~\ref{tab:tartan_dark} compares performance on the original and degraded observations.
When the RGB observations are degraded, the performance of the RGB-dependent methods deteriorates substantially, whereas DAPEVO maintains an AUC of $0.63$ and an ATE of $0.60$m.
It also improves over event-only DEVO, indicating that the remaining RGB information continues to provide useful correspondence constraints despite the degradation.
These results support the use of adaptive correlation fusion rather than committing to either modality globally.

\textbf{Reduced RGB frame rate.}
We next evaluate robustness to temporally sparse RGB observations by progressively subsampling the image stream while retaining event measurements between consecutive frames.
On TartanEvent, the RGB rate is reduced from $30$ to $3$ Hz, while DAPEVO processes event voxel grids at a fixed rate of $30$ Hz.
As shown in Table~\ref{tab:tartan_multirate}, all methods perform similarly at the nominal frame rate, but their behavior diverges as RGB observations become less frequent.
At $6$ Hz, DAPEVO obtains an ATE of $0.31$ m, compared with $1.70$ m for DPVO and $3.33$ m for RAMP-VO.
At $3$ Hz, DAPEVO remains below $1$ m ATE ($0.82$ m), whereas DPVO and RAMP-VO reach $9.58$ m and $9.97$ m, respectively.
The corresponding AUC remains $0.58$ for DAPEVO, compared with $0.09$ and $0.08$ for DPVO and RAMP-VO.

We further evaluate robustness to reduced RGB frame rates on the real-world UZH-FPV sequences.
Figure~\ref{fig:uzh_fpv_ate_drop} reports the mean ATE as the proportion of dropped RGB frames increases from $0$ to $83.3\%$.
\begin{figure}[!t]
    \centering
    \includegraphics[width=\columnwidth]{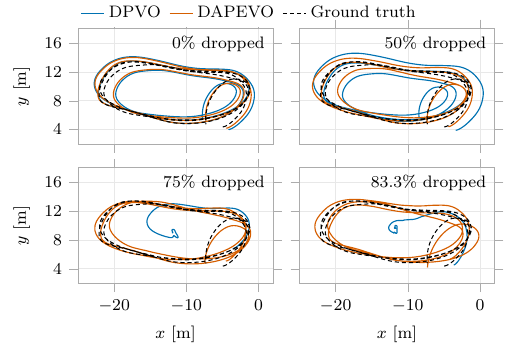}
    \caption{X--Y trajectories on UZH-FPV \texttt{indoor\_45\_13} at increasing image-drop rates.
As the drop rate increases, DPVO exhibits substantial trajectory deviation, whereas DAPEVO remains closely aligned with the ground truth.}
    \label{fig:uzh_fpv_multi_rate}
\end{figure}

DAPEVO's mean ATE increases from $1.00$ to $1.36$ m, compared with increases from $1.25$ to $4.63$ m for DPVO and from $1.49$ to $4.62$ m for RAMP-VO.
Figure~\ref{fig:uzh_fpv_multi_rate} compares trajectories on
\texttt{indoor\_45\_13} at RGB frame-drop rates of $0$, $50$, $75$, and $83.3\%$.
DAPEVO follows the ground-truth trajectory more closely, whereas DPVO exhibits substantial trajectory deviations.
Together, these results support the benefit of incorporating intermediate event-only observations when RGB measurements become temporally sparse.

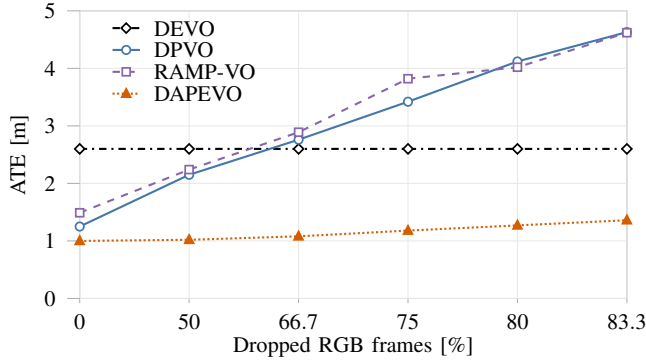
\begin{figure}[t]
    \centering
  \definecolor{plotDPVO}{HTML}{4477AA}
  \definecolor{plotDPEVO}{HTML}{D55E00}
  \definecolor{plotRAMP}{HTML}{8866AA}
    \resizebox{\columnwidth}{!}{%
    \begin{tikzpicture}[
        x=0.078cm,
        y=0.82cm,
        font=\small,
        devo/.style={
            black,
            line width=0.85pt,
            dash dot
        },
        dpvo/.style={
            plotDPVO,
            line width=0.85pt,
            solid
        },
        ramp/.style={
            plotRAMP,
            line width=0.85pt,
            dashed
        },
        dpevo/.style={
            plotDPEVO,
            line width=0.85pt,
            densely dotted
        }
    ]

    \draw[gray!65] (0,0) rectangle (100,5);

    \foreach \x/\label in {
        0/$0$,
        20/$50$,
        40/$66.7$,
        60/$75$,
        80/$80$,
        100/$83.3$
    } {
        \draw[gray!18,line width=0.3pt] (\x,0) -- (\x,5);
        \draw[gray!65]
            (\x,0) -- ++(0,-0.08);

        \node[anchor=north]
            at (\x,-0.12) {\label};
    }

    \foreach \ate/\ypos in {
        0/0,
        1/1,
        2/2,
        3/3,
        4/4,
        5/5
    } {
        \draw[gray!18,line width=0.3pt] (0,\ypos) -- (100,\ypos);
        \draw[gray!65]
            (0,\ypos) -- ++(-2.3,0);

        \node[anchor=east]
            at (-2.8,\ypos) {\ate};
    }

    \node at (50,-0.85) {Dropped RGB frames [\%]};
    \node[rotate=90] at (-11,2.5) {ATE [m]};

    \draw[devo]
        (0,2.60) -- (100,2.60);

    \foreach \x in {0,20,40,60,80,100} {
        \draw[
            black,
            line width=0.7pt,
            fill=white
        ]
            ($(\x,2.60)+(0,2.0pt)$) --
            ($(\x,2.60)+(-2.0pt,0)$) --
            ($(\x,2.60)+(0,-2.0pt)$) --
            ($(\x,2.60)+(2.0pt,0)$) --
            cycle;
    }

    \draw[dpvo]
        (0,1.25) --
        (20,2.15) --
        (40,2.76) --
        (60,3.42) --
        (80,4.12) --
        (100,4.63);

    \foreach \x/\y in {
        0/1.25,
        20/2.15,
        40/2.76,
        60/3.42,
        80/4.12,
        100/4.63
    } {
        \draw[
            plotDPVO,
            line width=0.7pt,
            fill=white
        ]
            (\x,\y) circle[radius=1.7pt];
    }

    \draw[ramp]
        (0,1.49) --
        (20,2.24) --
        (40,2.89) --
        (60,3.82) --
        (80,4.02) --
        (100,4.62);

    \foreach \x/\y in {
        0/1.49,
        20/2.24,
        40/2.89,
        60/3.82,
        80/4.02,
        100/4.62
    } {
        \draw[
            plotRAMP,
            line width=0.7pt,
            fill=white
        ]
            ($(\x,\y)+(-1.6pt,-1.6pt)$)
            rectangle
            ($(\x,\y)+(1.6pt,1.6pt)$);
    }

    \draw[dpevo]
        (0,1.00) --
        (20,1.02) --
        (40,1.08) --
        (60,1.18) --
        (80,1.27) --
        (100,1.36);

    \foreach \x/\y in {
        0/1.00,
        20/1.02,
        40/1.08,
        60/1.18,
        80/1.27,
        100/1.36
    } {
        \draw[
            plotDPEVO,
            line width=0.7pt,
            fill=plotDPEVO
        ]
            ($(\x,\y)+(0,2.0pt)$) --
            ($(\x,\y)+(-1.8pt,-1.4pt)$) --
            ($(\x,\y)+(1.8pt,-1.4pt)$) --
            cycle;
    }

    \begin{scope}[shift={(6,4.70)}]
        \draw[devo] (0,0) -- (5,0);
        \draw[black,line width=0.7pt,fill=white]
            ($(2.5,0)+(0,2pt)$) -- ($(2.5,0)+(-2pt,0)$) --
            ($(2.5,0)+(0,-2pt)$) -- ($(2.5,0)+(2pt,0)$) -- cycle;
        \node[anchor=west] at (6,0) {DEVO};

        \draw[dpvo] (0,-0.38) -- (5,-0.38);
        \draw[plotDPVO,line width=0.7pt,fill=white]
            (2.5,-0.38) circle[radius=1.7pt];
        \node[anchor=west] at (6,-0.38) {DPVO};

        \draw[ramp] (0,-0.76) -- (5,-0.76);
        \draw[plotRAMP,line width=0.7pt,fill=white]
            ($(2.5,-0.76)+(-1.6pt,-1.6pt)$) rectangle
            ($(2.5,-0.76)+(1.6pt,1.6pt)$);
        \node[anchor=west] at (6,-0.76) {RAMP-VO};

        \draw[dpevo] (0,-1.14) -- (5,-1.14);
        \draw[plotDPEVO,line width=0.7pt,fill=plotDPEVO]
            ($(2.5,-1.14)+(0,2pt)$) -- ($(2.5,-1.14)+(-1.8pt,-1.4pt)$) --
            ($(2.5,-1.14)+(1.8pt,-1.4pt)$) -- cycle;
        \node[anchor=west] at (6,-1.14) {DAPEVO};
    \end{scope}

    \end{tikzpicture}%
    }

\caption{UZH-FPV ATE averaged over five trials per sequence and all sequences. Equally spaced markers denote retaining every $k$th RGB frame, $k=1,\ldots,6$; labels give the corresponding drop percentages.}
    \label{fig:uzh_fpv_ate_drop}
\end{figure}

\begin{table}[t]
    \centering
    \caption{
        TartanEvent validation AUC and ATE [m] at reduced RGB rates.
        DAPEVO event voxel grids remain at 30\,Hz.
    }
    \label{tab:tartan_multirate}
    \scriptsize
    \setlength{\tabcolsep}{2.0pt}
    \begin{tabular}{@{}ccccccc@{}}
        \toprule
        RGB rate
        & \multicolumn{2}{c}{DPVO~\cite{teed2023deep}}
        & \multicolumn{2}{c}{RAMP-VO~\cite{pellerito2024rampvo}}
        & \multicolumn{2}{c}{DAPEVO} \\
        \cmidrule(lr){2-3}
        \cmidrule(lr){4-5}
        \cmidrule(l){6-7}
        [Hz]
        & AUC $\uparrow$
        & ATE $\downarrow$
        & AUC $\uparrow$
        & ATE $\downarrow$
        & AUC $\uparrow$
        & ATE $\downarrow$ \\
        \midrule

        30
        & 0.83 & 0.17
        & \textbf{0.85} & \textbf{0.15}
        & \textbf{0.85} & \textbf{0.15} \\

        15
        & \textbf{0.80} & 0.39
        & 0.74 & 0.45
        & 0.79 & \textbf{0.21} \\

        10
        & \textbf{0.77} & 0.45
        & 0.66 & 1.32
        & \textbf{0.77} & \textbf{0.29} \\

        6
        & 0.48 & 1.70
        & 0.39 & 3.33
        & \textbf{0.73} & \textbf{0.31} \\

        5
        & 0.34 & 3.64
        & 0.27 & 4.45
        & \textbf{0.69} & \textbf{0.35} \\

        3
        & 0.09 & 9.58
        & 0.08 & 9.97
        & \textbf{0.58} & \textbf{0.82} \\

        \bottomrule
    \end{tabular}
\end{table}

\subsection{Ablation studies}
\label{sec:ablation}
\begin{figure}[!t]
  \centering
  \includegraphics[width=\columnwidth]{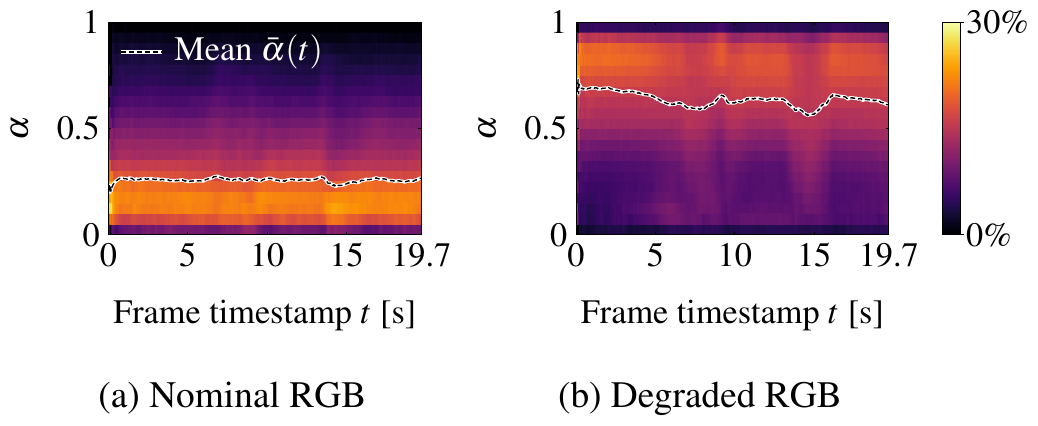}
  \caption{Temporal distribution of the event-fusion weight $\alpha$ on
  TartanEvent \texttt{japanesealley/Hard/P005}. Columns show normalized edges fractions in $0.05$-wide bins; we exclude edges whose reprojections fall out-of-frame. Dashed curves show the mean. 
  Larger $\alpha$ favors events, while smaller $\alpha$ favors RGB.}
  \label{fig:alpha-distribution-tro}
\end{figure}

\textbf{Adaptive Correlation Fusion}
We first assess whether observation-dependent fusion improves performance beyond a fixed modality weighting.
Figure~\ref{fig:alpha-distribution-tro} shows the distribution of the adaptive fusion weight $\alpha_{kj}$ for the same sequence under nominal and degraded RGB input.

The mean event-fusion weight increases from approximately $0.25$ under nominal conditions to $0.6$-$0.7$ under degradation, indicating greater reliance on event correspondences as image quality deteriorates.
This raises the question of whether the benefit comes from adapting the weight to individual observations or simply shifting the overall balance between modalities.

To distinguish these effects, we replace $\alpha_{kj}$ with four alternatives: $0$ (RGB-only correlations), $1$ (event-only correlations), $0.5$ (equal weighting), and a condition-specific constant equal to the mean weight predicted by the adaptive model, denoted \textit{Dataset Mean}.
The latter tests whether a global modality balance derived from the adaptive model is sufficient to recover its performance.
All variants are evaluated on the TartanEvent validation split under both nominal and degraded RGB input.

Table~\ref{tab:fixed_gate} shows that under nominal conditions, RGB-only matching achieves the lowest ATE, while adaptive fusion remains close in both ATE and AUC.
Under degraded RGB input, adaptive fusion achieves an ATE of $0.60$ m and an AUC of $0.63$, compared with $0.73$ m and $0.59$ for event-only matching, corresponding to a $17.8\%$ reduction in ATE.
The Dataset Mean variant yields an ATE of $0.99$ m, showing that replacing observation-dependent weights with their condition-specific average does not preserve the benefit.
Together, these results support adapting the fusion weight across observations to retain useful image information while increasing reliance on events when image correspondences become unreliable.

\begin{table}[t]
\centering
\caption{Fusion-weight ablation on TartanEvent validation. \normalfont\itshape{Dataset Mean} fixes $\alpha$ to the adaptive model's mean over the considered dataset.}
\label{tab:fixed_gate}
\scriptsize
\setlength{\tabcolsep}{5pt}
\begin{tabular}{c|cc|cc}
\toprule
& \multicolumn{2}{c|}{Nominal RGB}
& \multicolumn{2}{c}{Degraded RGB} \\
$\alpha$
& AUC $\uparrow$ & ATE [m] $\downarrow$
& AUC $\uparrow$ & ATE [m] $\downarrow$ \\
\midrule
0.0 (RGB only)
& \textbf{0.87} & \textbf{0.13}
& 0.19 & 5.04 \\

Dataset Mean
& 0.84 & 0.16
& 0.57 & 0.99 \\

0.5
& 0.82 & 0.19
& 0.57 & 0.99 \\

1.0 (events only)
& 0.58 & 0.96
& 0.59 & 0.73 \\

Adaptive
& 0.85 & 0.15
& \textbf{0.63} & \textbf{0.60} \\
\bottomrule
\end{tabular}
\end{table}

\textbf{Keyframe management.}
We next evaluate the proposed modality-aware keyframe management against the original culling strategy used by DPVO and DEVO.
The comparison is performed on the TartanEvent validation split with the RGB frame rate reduced to 5 Hz, such that joint RGB-event observations are substantially less frequent than event-only observations.
Table~\ref{tab:culling_ablation} compares the original and proposed subgraph-based strategies using equal culling thresholds for both observation types, $(15,15)$, and the modality-specific thresholds $(15,25)$ used in DAPEVO.
The original strategy retains only $20\%$ joint keyframes with equal thresholds, whereas the proposed strategy increases this fraction to $54\%$.
With modality-specific thresholds, the fraction of retained joint keyframes further increases to $74\%$.
This preservation of frame-bearing keyframes translates into improved trajectory accuracy: the proposed strategy with $(15,25)$ increases AUC from $0.64$ to $0.69$ and reduces ATE from $0.71$ m to $0.35$ m relative to the corresponding original culling strategy.

\begin{table}[t]
\centering
\caption{
Culling ablation on TartanEvent validation at 5 Hz RGB.
Joint frames: fraction of retained keyframes with both modalities.
}
\label{tab:culling_ablation}
\scriptsize
\setlength{\tabcolsep}{3.0pt}
\begin{tabular}{@{}lcccc@{}}
\toprule
Method
& \shortstack{Threshold\\(joint, evt-only)}
& \shortstack{AUC\\$\uparrow$}
& \shortstack{ATE [m]\\$\downarrow$}
& \shortstack{Joint\\frames} \\
\midrule

    Original~\cite{teed2023deep,klenk2024devo}
    & (15, 15)
    & 0.57
    & 0.70
    & 0.20 \\

    Original~\cite{teed2023deep,klenk2024devo}
    & (15, 25)
    & 0.64
    & 0.71
    & 0.57 \\

    Subgraph
    & (15, 15)
    & 0.66
    & 0.39
    & 0.54 \\

    Subgraph
    & (15, 25)
    & \textbf{0.69}
    & \textbf{0.35}
    & 0.74 \\

    \bottomrule
\end{tabular}

\end{table}

\textbf{Patch selection strategy.}
DAPEVO uses the event-based patch scorer introduced in DEVO to select patch locations shared by both modalities.
Because DPVO reports random patch sampling as its strongest selection strategy, we verify that replacing it with the DEVO event scorer does not degrade the underlying frame-based pipeline.
We evaluate DPVO with the DEVO event scorer, and find that the event-based patch selection does not degrade RGB frame-only tracking; instead, AUC increases from $0.83$ to $0.86$ and ATE decreases from $0.17$ m to $0.14$ m.
Consequently, the use of the event scorer is not responsible for a loss in the underlying frame-based performance and provides a suitable common patch-selection mechanism for the two modalities.

\section{Discussion and Conclusion}
\label{sec:conclusion}
We presented DAPEVO, an image-event visual odometry method that fuses modality-specific evidence at the patch-correlation level while maintaining a shared geometric state.
Event-only observations are incorporated into the same optimization graph, while modality-aware keyframe management preserves joint observations required for frame-based correlations.
Experiments under reduced RGB frame rates show that intermediate event observations substantially improve robustness as frame observations become sparse, on both synthetic and real-world data.
Under challenging illumination, adaptive correlation fusion reduces the dependence on unreliable RGB frame information while retaining useful visual constraints, outperforming either modality alone.
Together, these results demonstrate the benefit of performing image-event fusion at the correspondence level and adapting the contribution of each modality to the available observations.

\section*{Acknowledgment}
OpenAI Codex (GPT-6) assisted with generating Figure 1 and editing the manuscript for clarity and readability. Claude Opus 5 assisted with generating Figure 2.


\begin{thebibliography}{10}
\providecommand{\url}[1]{#1}
\csname url@samestyle\endcsname
\providecommand{\newblock}{\relax}
\providecommand{\bibinfo}[2]{#2}
\providecommand{\BIBentrySTDinterwordspacing}{\spaceskip=0pt\relax}
\providecommand{\BIBentryALTinterwordstretchfactor}{4}
\providecommand{\BIBentryALTinterwordspacing}{\spaceskip=\fontdimen2\font plus
\BIBentryALTinterwordstretchfactor\fontdimen3\font minus
  \fontdimen4\font\relax}
\providecommand{\BIBforeignlanguage}[2]{{%
\expandafter\ifx\csname l@#1\endcsname\relax
\typeout{** WARNING: IEEEtran.bst: No hyphenation pattern has been}%
\typeout{** loaded for the language `#1'. Using the pattern for}%
\typeout{** the default language instead.}%
\else
\language=\csname l@#1\endcsname
\fi
#2}}
\providecommand{\BIBdecl}{\relax}
\BIBdecl

\bibitem{carlone2026slam}
L.~Carlone, A.~Kim, T.~D. Barfoot, D.~Cremers, and F.~Dellaert, Eds.,
  \emph{{SLAM Handbook: From Localization and Mapping to Spatial
  Intelligence}}.\hskip 1em plus 0.5em minus 0.4em\relax Cambridge University
  Press, 2026.

\bibitem{tranzatto2024team}
M.~Tranzatto, M.~Dharmadhikari, L.~Bernreiter, M.~Camurri, S.~Khattak,
  F.~Mascarich, P.~Pfreundschuh, D.~Wisth, S.~Zimmermann, M.~Kulkarni
  \emph{et~al.}, ``Team cerberus wins the darpa subterranean challenge:
  Technical overview and lessons learned,'' \emph{Field Robotics}, vol.~4, pp.
  349--312, 2024.

\bibitem{zhao2024subt}
S.~Zhao, Y.~Gao, T.~Wu, D.~Singh, R.~Jiang, H.~Sun, M.~Sarawata, Y.~Qiu,
  W.~Whittaker, I.~Higgins \emph{et~al.}, ``Subt-mrs dataset: Pushing slam
  towards all-weather environments,'' in \emph{Proceedings of the IEEE/CVF
  conference on computer vision and pattern recognition}, 2024, pp.
  22\,647--22\,657.

\bibitem{cheng2005visual}
Y.~Cheng, M.~Maimone, and L.~Matthies, ``Visual odometry on the mars
  exploration rovers,'' in \emph{2005 IEEE International Conference on Systems,
  Man and Cybernetics}, vol.~1.\hskip 1em plus 0.5em minus 0.4em\relax IEEE,
  2005, pp. 903--910.

\bibitem{gallego2022survey}
G.~Gallego, T.~Delbr{\"u}ck, G.~Orchard, C.~Bartolozzi, B.~Taba, A.~Censi,
  S.~Leutenegger, A.~J. Davison, J.~Conradt, K.~Daniilidis, and D.~Scaramuzza,
  ``Event-based vision: A survey,'' \emph{IEEE Trans. Pattern Anal. Mach.
  Intell.}, vol.~44, no.~1, pp. 154--180, 2022.

\bibitem{teed2023deep}
Z.~Teed, L.~Lipson, and J.~Deng, ``{Deep Patch Visual Odometry},'' \emph{Adv.
  Neural Inf. Process. Syst.}, vol.~36, pp. 39\,033--39\,051, 2023.

\bibitem{klenk2024devo}
S.~Klenk, M.~Motzet, L.~Koestler, and D.~Cremers, ``{Deep Event Visual
  Odometry},'' in \emph{Proc. Int. Conf. 3D Vis. (3DV)}, 2024, pp. 739--749.

\bibitem{hidalgo2022eds}
J.~Hidalgo-Carri{\'o}, G.~Gallego, and D.~Scaramuzza, ``Event-aided direct
  sparse odometry,'' in \emph{Proc. IEEE/CVF Conf. Comput. Vis. Pattern
  Recognit. (CVPR)}, 2022, pp. 5781--5790.

\bibitem{pellerito2024rampvo}
R.~Pellerito, M.~Cannici, D.~Gehrig, J.~Belhadj, O.~Dubois-Matra, M.~Casasco,
  and D.~Scaramuzza, ``{Deep Visual Odometry with Events and Frames},'' in
  \emph{Proc. IEEE/RSJ Int. Conf. Intell. Robots Syst. (IROS)}, 2024, pp.
  8966--8973.

\bibitem{delmerico2019uzh}
J.~Delmerico, T.~Cieslewski, H.~Rebecq, M.~Faessler, and D.~Scaramuzza, ``Are
  we ready for autonomous drone racing? the {UZH-FPV} drone racing dataset,''
  in \emph{Proc. IEEE Int. Conf. Robot. Autom. (ICRA)}, 2019.

\bibitem{campos2021orbslam3}
C.~Campos, R.~Elvira, J.~J. G{\'o}mez~Rodr{\'i}guez, J.~M.~M. Montiel, and
  J.~D. Tard{\'o}s, ``{ORB-SLAM3}: An accurate open-source library for visual,
  visual-inertial, and multi-map {SLAM},'' \emph{IEEE Trans. Robot.}, vol.~37,
  no.~6, pp. 1874--1890, 2021.

\bibitem{engel2018dso}
J.~Engel, V.~Koltun, and D.~Cremers, ``Direct sparse odometry,'' \emph{IEEE
  Trans. Pattern Anal. Mach. Intell.}, vol.~40, no.~3, pp. 611--625, 2018.

\bibitem{wang2017deepvo}
S.~Wang, R.~Clark, H.~Wen, and N.~Trigoni, ``{DeepVO}: Towards end-to-end
  visual odometry with deep recurrent convolutional neural networks,'' in
  \emph{Proc. IEEE Int. Conf. Robot. Autom. (ICRA)}, 2017, pp. 2043--2050.

\bibitem{yang2020d3vo}
N.~Yang, L.~von Stumberg, R.~Wang, and D.~Cremers, ``{D3VO}: Deep depth, deep
  pose and deep uncertainty for monocular visual odometry,'' in \emph{Proc.
  IEEE/CVF Conf. Comput. Vis. Pattern Recognit. (CVPR)}, 2020, pp. 1281--1292.

\bibitem{teed2020raft}
Z.~Teed and J.~Deng, ``{RAFT}: Recurrent all-pairs field transforms for optical
  flow,'' in \emph{Proc. Eur. Conf. Comput. Vis. (ECCV)}, 2020, pp. 402--419.

\bibitem{teed2021droid}
------, ``{DROID-SLAM: Deep Visual SLAM for Monocular, Stereo, and RGB-D
  Cameras},'' \emph{Adv. Neural Inf. Process. Syst.}, vol.~34, 2021.

\bibitem{lipson2024deep}
L.~Lipson, Z.~Teed, and J.~Deng, ``{Deep Patch Visual SLAM},'' in \emph{Proc.
  Eur. Conf. Comput. Vis. (ECCV)}, 2024, pp. 424--440.

\bibitem{kim2016tracking}
H.~Kim, S.~Leutenegger, and A.~J. Davison, ``Real-time {3D} reconstruction and
  {6-DoF} tracking with an event camera,'' in \emph{Proc. Eur. Conf. Comput.
  Vis. (ECCV)}, 2016, pp. 349--364.

\bibitem{rebecq2017evo}
H.~Rebecq, T.~Horstschaefer, G.~Gallego, and D.~Scaramuzza, ``{EVO}: A
  geometric approach to event-based {6-DOF} parallel tracking and mapping in
  real time,'' \emph{IEEE Robot. Autom. Lett.}, vol.~2, no.~2, pp. 593--600,
  2017.

\bibitem{zhou2021esvo}
Y.~Zhou, G.~Gallego, and S.~Shen, ``Event-based stereo visual odometry,''
  \emph{IEEE Trans. Robot.}, vol.~37, no.~5, pp. 1433--1450, 2021.

\bibitem{zhu2017evio}
A.~Z. Zhu, N.~Atanasov, and K.~Daniilidis, ``Event-based visual inertial
  odometry,'' in \emph{Proc. IEEE Conf. Comput. Vis. Pattern Recognit. (CVPR)},
  2017, pp. 5391--5399.

\bibitem{rebecq2017eventvio}
H.~Rebecq, T.~Horstschaefer, and D.~Scaramuzza, ``Real-time visual-inertial
  odometry for event cameras using keyframe-based nonlinear optimization,'' in
  \emph{Proc. Brit. Mach. Vis. Conf. (BMVC)}, 2017, pp. 16.1--16.12.

\bibitem{zhu2019unsupervised}
A.~Z. Zhu, L.~Yuan, K.~Chaney, and K.~Daniilidis, ``Unsupervised event-based
  learning of optical flow, depth, and egomotion,'' in \emph{Proc. IEEE/CVF
  Conf. Comput. Vis. Pattern Recognit. (CVPR)}, 2019, pp. 989--997.

\bibitem{deio}
W.~Guan, F.~Lin, P.~Chen, and P.~Lu, ``{DEIO: Deep Event Inertial Odometry},''
  in \emph{Proc. IEEE/CVF Int. Conf. Comput. Vis. Workshops (ICCVW)}, 2025, pp.
  4665--4674.

\bibitem{stereodevo}
S.~Zhong, J.~Niu, and Y.~Zhou, ``{Deep Visual Odometry for Stereo Event
  Cameras},'' \emph{IEEE Robot. Autom. Lett.}, vol.~10, no.~11, pp.
  11\,078--11\,085, 2025.

\bibitem{kueng2016tracks}
B.~Kueng, E.~Mueggler, G.~Gallego, and D.~Scaramuzza, ``Low-latency visual
  odometry using event-based feature tracks,'' in \emph{Proc. IEEE/RSJ Int.
  Conf. Intell. Robots Syst. (IROS)}, 2016, pp. 16--23.

\bibitem{vidal2018ultimate}
A.~Rosinol~Vidal, H.~Rebecq, T.~Horstschaefer, and D.~Scaramuzza, ``{Ultimate
  SLAM? Combining Events, Images, and IMU for Robust Visual SLAM in HDR and
  High-Speed Scenarios},'' \emph{IEEE Robot. Autom. Lett.}, vol.~3, no.~2, pp.
  994--1001, 2018.

\bibitem{guan2023plevio}
W.~Guan, P.~Chen, Y.~Xie, and P.~Lu, ``{PL-EVIO: Robust Monocular Event-Based
  Visual Inertial Odometry With Point and Line Features},'' \emph{IEEE Trans.
  Autom. Sci. Eng.}, 2023.

\bibitem{gehrig2021ramnet}
D.~Gehrig, M.~R{\"u}egg, M.~Gehrig, J.~Hidalgo-Carri{\'o}, and D.~Scaramuzza,
  ``Combining events and frames using recurrent asynchronous multimodal
  networks for monocular depth prediction,'' \emph{IEEE Robot. Autom. Lett.},
  vol.~6, no.~2, pp. 2822--2829, 2021.

\bibitem{tartanair}
W.~Wang, D.~Zhu, X.~Wang, Y.~Hu, Y.~Qiu, C.~Wang, Y.~Hu, A.~Kapoor, and
  S.~Scherer, ``{TartanAir: A Dataset to Push the Limits of Visual SLAM},'' in
  \emph{Proc. IEEE/RSJ Int. Conf. Intell. Robots Syst. (IROS)}, 2020.

\bibitem{guan2022monocular}
W.~Guan and P.~Lu, ``Monocular event visual inertial odometry based on
  event-corner using sliding windows graph-based optimization,'' in \emph{2022
  IEEE/RSJ International Conference on Intelligent Robots and Systems
  (IROS)}.\hskip 1em plus 0.5em minus 0.4em\relax IEEE, 2022, pp. 2438--2445.

\bibitem{zhu2018multivehicle}
A.~Z. Zhu, D.~Thakur, T.~{\"O}zaslan, B.~Pfrommer, V.~Kumar, and K.~Daniilidis,
  ``The multi vehicle stereo event camera dataset: An event camera dataset for
  3d perception,'' \emph{IEEE Robotics and Automation Letters}, vol.~3, no.~3,
  pp. 2032--2039, 2018.

\bibitem{qin2018vinsmono}
T.~Qin, P.~Li, and S.~Shen, ``{VINS-Mono}: A robust and versatile monocular
  visual-inertial state estimator,'' \emph{IEEE Trans. Robot.}, vol.~34, no.~4,
  pp. 1004--1020, 2018.

\bibitem{qin2019vinsfusion}
T.~Qin, S.~Cao, J.~Pan, and S.~Shen, ``A general optimization-based framework
  for global pose estimation with multiple sensors,'' \emph{arXiv preprint
  arXiv:1901.03642}, 2019.

\bibitem{chen2023esvio}
P.~Chen, W.~Guan, and P.~Lu, ``Esvio: Event-based stereo visual inertial
  odometry,'' \emph{IEEE Robotics and Automation Letters}, vol.~8, no.~6, pp.
  3661--3668, 2023.

\bibitem{schoenberger2016sfm}
J.~L. Sch{\"o}nberger and J.-M. Frahm, ``Structure-from-motion revisited,'' in
  \emph{Proc. IEEE Conf. Comput. Vis. Pattern Recognit. (CVPR)}, 2016.

\end{thebibliography}
\end{document}